# AN ADAPTIVE DATA-CLEANING FRAMEWORK FOR NOISY-LABEL DETECTION

**Chen-Hsuan Fang[a], Wei-Hsiang Chen[a], Pin-Hsuan Yu[a], Jung-Hua Wang[a,b*], and Tsung-Wei Pan[a]**

[a] Department of Electrical Eng. National Taiwan Ocean University, Keelung City 20224, Taiwan
[b] AI Research Center, National Taiwan Ocean University, Keelung City 20224, Taiwan
[*] Corresponding author: Jung-Hua Wang (e-mail: jhwang@email.ntou.edu.tw).

## Abstract

Deep neural networks (DNNs) have demonstrated exceptional performance across computer vision tasks owing to large-scale annotated datasets. In real-world applications, however, labels are corrupted by annotation ambiguity, human error, or dynamically changing environments. Over-parameterized DNNs memorize these noisy labels during training, resulting in degradation of model accuracy and generalization. Existing data-cleaning and sample-selection strategies address this challenge, yet often rely on manually specified thresholds, prior knowledge of the noise ratio, or a single metric, either learning dynamics or geometric structure, rendering them unstable in complex data regimes. This paper proposes a self-adaptive data-cleaning framework that integrates local, global, and learning dynamics cues for robust noisy-label detection. Samples are mapped into a unified low-dimensional feature space through a modular feature concatenation paradigm. This paradigm provides two instantiations: a 2D metric integrating class-adaptive KNN-based local disagreement and $k$-means-based global centroid distance, and a 3D multi-metric that additionally incorporates a well-defined z-normalized score. Unlike conventional 1D Gaussian Mixture Models applied to a single scalar metric, the proposed framework performs multi-metric clustering on the feature space and adaptively partitions samples into clean-dominant and noise-dominant components without requiring manual filtering thresholds or noise-ratio priors. Experiments on CIFAR-10, MNIST, and ImageNet-100 with 5% to 40% symmetric label noise show high recall in most settings, including near-perfect recall (≥98%) on ImageNet-100 at 40% noise. Subsequent model training further shows accuracy gains in many evaluated settings, especially under severe corruption on ImageNet-100. These findings suggest that multi-metric integration provides a threshold-free, practical, and low-tuning strategy for noisy-label detection.

# 1 Introduction

Deep neural networks (DNNs) have achieved remarkable performance across a wide range of computer vision and machine learning tasks, largely owing to the availability of large-scale annotated datasets [1]. In practice, however, training labels are often corrupted by annotation ambiguity, human error, weak supervision, or dynamically changing environments. As a result, label noise has become a pervasive issue in real-world datasets [2], [3]. Because overparameterized DNNs can readily memorize incorrect labels during training, label noise can severely degrade model accuracy, robustness, and generalization [4]. Developing effective strategies for learning with noisy labels is therefore essential for building reliable deep learning systems [5]. To rigorously evaluate the proposed framework, this study specifically focuses on symmetric label noise. Under this setting, a predefined percentage of true labels from each class are randomly sampled and uniformly reassigned to all other available classes, thereby simulating a controlled and unbiased noise distribution for evaluation.

To address this challenge, a substantial body of research has focused on noisy-label detection and data cleaning. Existing approaches can be broadly categorized into three directions: learning dynamics-based methods, geometric-based methods, and probabilistic or statistical frameworks [6], [7], [8]. Here, learning dynamics is defined as sample-level error- or loss-related metric observed during training, reflecting how easily or reliably individual samples are learned by the model. These studies have established useful foundations for identifying unreliable samples and reducing the adverse effects of label noise. Nevertheless, their practical effectiveness remains limited in many realistic settings.

Two challenges are particularly important. First, many existing methods rely on manually specified thresholds, filtering ratios, or prior assumptions about the noise rate [9]. Such settings are often highly dataset-dependent and difficult to transfer across different architectures, datasets, and corruption conditions. Second, many methods rely primarily on a single source of evidence, which can lead to unstable detection performance in complex data regimes [10]. Recent pruning studies have further suggested that sample difficulty alone is insufficient for robust sample selection, highlighting the need to integrate complementary evidence such as prediction uncertainty [11] rather than relying on any single criterion for noisy-label detection.

To overcome these limitations, this paper proposes a self-adaptive, multi-metric data-cleaning framework specifically designed for robust noisy-label detection. The central idea is to integrate complementary local, global, and learning dynamics related cues into a unified low-dimensional feature space. Specifically, the framework projects samples into either a 2D feature space, which encodes class-adaptive KNN local disagreement ratio and $k$-means-based global centroid distance [12], [13], [14] or a 3D feature space that further incorporates an error metric to capture learning dynamics [15]. A Gaussian Mixture Model (GMM) is subsequently fitted to this feature space to perform adaptive clean-noise partitioning [16],[17]. In contrast to existing methods that rely on isolated metrics or manually tuned cutoffs, this method is fully automated and threshold-free, eliminating the necessity for dataset-specific hyperparameters or prior knowledge of the noise ratio. By comprehensively multiple sources of evidence, the proposed approach establishes a highly transferable and practical solution for data refinement.

The main contributions of this paper are fourfold. First, we propose a threshold-free adaptive partitioning mechanism based on multi-dimensional GMM clustering, which identifies the noise-dominant component without requiring manually specified filtering thresholds or prior knowledge of the noise ratio. Second, we introduce a unified multi-metric for noisy-label detection that integrates local disagreement ratio, global centroid distance, and learning dynamics information. Third, we develop a modular data-cleaning framework with two variants: a 2D geometric metric based on local disagreement ratio and global centroid distance, and a 3D multi-metric that further incorporates a robust EL2N score. Fourth, we conduct comprehensive empirical validation on CIFAR-10, MNIST, and ImageNet-100 using ResNet50 and Vision Transformer backbones, demonstrating high noisy-label recall and improved downstream retraining performance in many evaluated settings.

# 2 Related work

Learning with noisy labels has motivated a broad range of data-cleaning and sample-selection methods aimed at identifying unreliable samples and reducing the adverse effects of corrupted labels during model training. It is well documented that overparameterized DNNs possess a strong propensity to memorize corrupted labels, causing noisy annotations to severely degrade model accuracy and generalization [4]. To address this fundamental challenge, existing literature has predominantly focused on data pruning and sample selection mechanisms [5]. This section reviews the prominent approaches in this domain categorized into learning dynamics-based, geometric structure, and automated probabilistic frameworks and critically analyzes their inherent limitations.

## 2.1 Sample Selection via Learning Dynamics

A substantial body of work leverages the learning dynamics of DNNs, predicated on the well-established small-loss phenomenon: clean and structurally simple samples are typically optimized earlier than their mislabeled counterparts [9], [18]. Among these approaches, the Error L2-Norm (EL2N) has emerged as a prominent metric for quantifying the learning difficulty of individual samples during the training epochs [15]. While EL2N provides valuable insights into the relationship between learning dynamics and sample importance, it was not originally designed as a dedicated noisy-label detector. A critical limitation of using EL2N as an independent noise-filtering criterion is that structurally ambiguous but correctly labeled samples and genuinely mislabeled samples may both produce high EL2N scores during training. Consequently, applying a naive threshold solely to EL2N may discard informative hard samples together with corrupted ones, especially when the feature space contains overlapping or ambiguous class boundaries.

Recent pruning studies further suggest that example difficulty alone is insufficient for robust sample selection. In [11], example difficulty was combined with prediction uncertainty to demonstrate that these complementary metrics improve pruning robustness under challenging conditions, including label noise and image corruption. Although their objective is not direct noisy-label detection, this finding reinforces the view that EL2N-like difficulty scores should not be used as standalone criteria for label-noise refinement. In particular, relying solely on difficulty metrics risks conflating structural unreliability with learning difficulty, lacking the geometric context needed for adaptive data refinement.

### 2.2 Geometric Feature Methods

To circumvent the shortcomings of purely loss-based metrics, an alternative line of research exploits the topological and geometric structures of the learned deep feature space [19],[20]. These methodologies are anchored in the assumption that correctly labeled samples tend to map into dense, semantically coherent clusters, whereas mislabeled samples manifest as local or global outliers. For instance, K-Nearest Neighbors (KNN) approaches assess local neighborhood consistency by measuring label agreement between a sample and its adjacent neighbors[12],[21]. Concurrently, global structural deviations are captured via clustering algorithms, such as *k*-means, which construct class centroid distance and quantify the discrepancy between each sample and its assigned centroid [13],[14].

However, both KNN-based and *k*-means-based geometric metrics exhibit limited efficacy in unconstrained, realistic conditions. They depend heavily on manually calibrated hyperparameters such as neighborhood size k, distance thresholds, or fixed centroid definitions. This reliance renders them particularly brittle in the presence of severe class imbalance, multi-modal class distributions, or substantial interclass overlap.

### 2.3 Lacking Automated Probabilistic Frameworks

To formalize noise detection, probabilistic and statistical frameworks have been proposed, with Confident Learning serving as a representative paradigm [22]. These methods estimate the joint distribution of noisy and true labels to systematically identify label errors and quantify model uncertainty. Despite offering a statistically principled foundation, such frameworks exhibit critical practical vulnerabilities. They rely heavily on the fidelity of out-of-sample predicted probabilities and typically incur substantial computational overhead due to cross-validation requirements [22]. More crucially, their performance is bottlenecked by the assumption that the underlying DNN produces well-calibrated confidence scores. When optimized on corrupted data, model calibration inevitably deteriorates, leading to skewed joint-distribution estimates and consequently suboptimal label cleaning outcomes [23].

Recognizing these calibration issues, contemporary research has sought alternative validation strategies. For example, ReCoV was proposed as a parameter-free label-noise detection method that identifies noisy samples by measuring how frequently they contribute to inferior cross-validation results [24]. This approach avoids requiring prior knowledge of the noise ratio and provides a model-agnostic validation-based mechanism for noisy-label detection. However, repeated cross-validations can introduce substantial computational cost, especially in deep learning settings where iterative model training is expensive. These insights emphasize that while automated probabilistic frameworks represent a promising direction, their practical utility hinges on operating within a discriminative and computationally efficient feature space.

### 2.4 Research Gaps and Contributions of the Proposed Framework

The aforementioned limitations underscore that robust noisy-label detection necessitates the synergistic integration of complementary evidence, rather than reliance on a singular criterion. Our method directly addresses the critical gaps identified in the latest literature. To the best of our knowledge, no prior literature has systematically combined localized geometric consistency, global centroid distance, and learning dynamics metrics into a unified, fully adaptive feature space. The primary distinction of our work lies in the proposed modular multi-metric paradigm. We present two concrete instantiations within a unified end-to-end pipeline: a 2D geometric metric consisting of class-adaptive KNN-based local disagreement ratio and $k$-means -based global centroid distance, and a comprehensive 3D multi-metric that additionally incorporates a class-conditional, robust Z-normalized EL2N score.

Unlike existing probabilistic approaches that typically apply one-dimensional Gaussian Mixture Models (GMMs) solely to a single scalar metric, rely on model confidence calibration [22],[23], or require computationally heavy repeated cross-validations [24], our framework performs multi-dimensional GMM clustering directly on the feature space. This architectural design eradicates the need for dataset-specific cutoff engineering and prior assumptions regarding noise ratios. Instead of predefining explicit scenarios for their application, these two variants are evaluated under the same experimental protocol to examine how geometric and learning dynamics metrics contribute to noisy-label data refinement across different datasets and backbone architectures. By validating both variants within the same empirical framework, the proposed approach demonstrates superior generalizability and establishes a robust, highly automated foundation for noisy-label data refinement. Based on the above research gaps and methodological distinctions, the main contributions of this study are summarized as follows

#### 2.4.1 Threshold-Free Adaptive Partitioning:

This study introduces a threshold-free adaptive partitioning mechanism based on multi-dimensional Gaussian Mixture Model clustering. Unlike existing approaches that require manually specified filtering thresholds, prior knowledge of the noise ratio, calibrated confidence scores, or repeated cross-validation, the proposed framework automatically identifies the noise-dominant component from the empirical distribution of the fused feature space. This mechanism reduces dataset-specific tuning and improves the practicality of noisy-label detection across different datasets, architectures, and corruption levels.

#### 2.4.2 Multi-Metric for Noisy-Label Detection:

This study proposes a unified multi-metric that integrates local disagreement ratio, global centroid distance, and Robust EL2N score for noisy-label detection. Instead of relying on a single indicator, such as loss values, neighborhood agreement, centroid distance, or model confidence alone, the proposed combines complementary evidence from different perspectives. This design enables the framework to distinguish noisy samples more robustly in complex feature spaces where individual metrics may be insufficient or unstable

#### 2.4.3 Modular Data-Cleaning Framework:

This study develops two modular variants within the same adaptive data-cleaning pipeline. The 2D geometric metric combines class-adaptive KNN-based local disagreement ratio with *k*-means-based global centroid distance to capture both local and global structural inconsistency. The 3D multi-metric further incorporates a class-conditional robust Z-normalized EL2N score to include learning dynamics information. By evaluating both variants under the same protocol, this study systematically investigates how geometric and learning dynamics cues contribute to noisy-label data refinement.

### 2.4.4 Comprehensive Empirical Validation:

This study provides extensive empirical validation on CIFAR-10, MNIST, and ImageNet-100 under multiple symmetric label-noise ratios and two representative backbone architectures, including ResNet50 and Vision Transformer. The experimental results show that the proposed framework achieves consistently high noisy-label recall across most evaluated settings and improves downstream retraining performance in many cases, particularly under severe corruption and complex feature-space conditions. These findings demonstrate the effectiveness and generalizability of the proposed adaptive data-cleaning strategy.

# 3 Methodology

The proposed framework aims to partition a training dataset into clean and noisy subsets through a low-tuning, self-adaptive pipeline. It integrates three complementary sample-level metrics: local disagreement ratio, global centroid distance, and robust EL2N score. The local disagreement ratio and global centroid distance are computed from feature embeddings extracted from the layer immediately before the classification layer of a trained backbone. In contrast, the EL2N score is derived from learning dynamics, where each sample obtains an EL2N score at every training epoch, and the epoch-wise EL2N scores of each sample are averaged across all training epochs. The averaged EL2N score is then transformed by a class-conditional robust Z-score normalization to obtain the robust EL2N score used in the 3D metric. These metrics are organized into either a 2D geometric metric or a 3D multi-metric representation, followed by an automated, threshold-free partitioning mechanism based on a multi-dimensional Gaussian Mixture Model (GMM).

Fig. 1 illustrates the overall workflow of the proposed adaptive data-cleaning framework. Given an input dataset, a backbone model is first trained on the noisy training set. During training, an EL2N score is computed for each sample at each epoch according to the discrepancy between the predicted probability vector and the observed label. After training, the epoch-wise EL2N scores of each sample are averaged and subsequently converted into a robust EL2N score through class-conditional robust normalization. Meanwhile, all samples are forwarded through the trained model, and feature embeddings are extracted from the layer immediately before the classification layer. Based on these embeddings, the local disagreement ratio and global centroid distance are computed for each sample. The 2D variant represents each sample using the local disagreement ratio and global centroid distance, whereas the 3D variant further incorporates the robust EL2N score. Finally, a two-component GMM is applied to the corresponding 2D or 3D metric space to adaptively partition samples into clean and removed subsets without requiring a manually specified cleaning threshold.

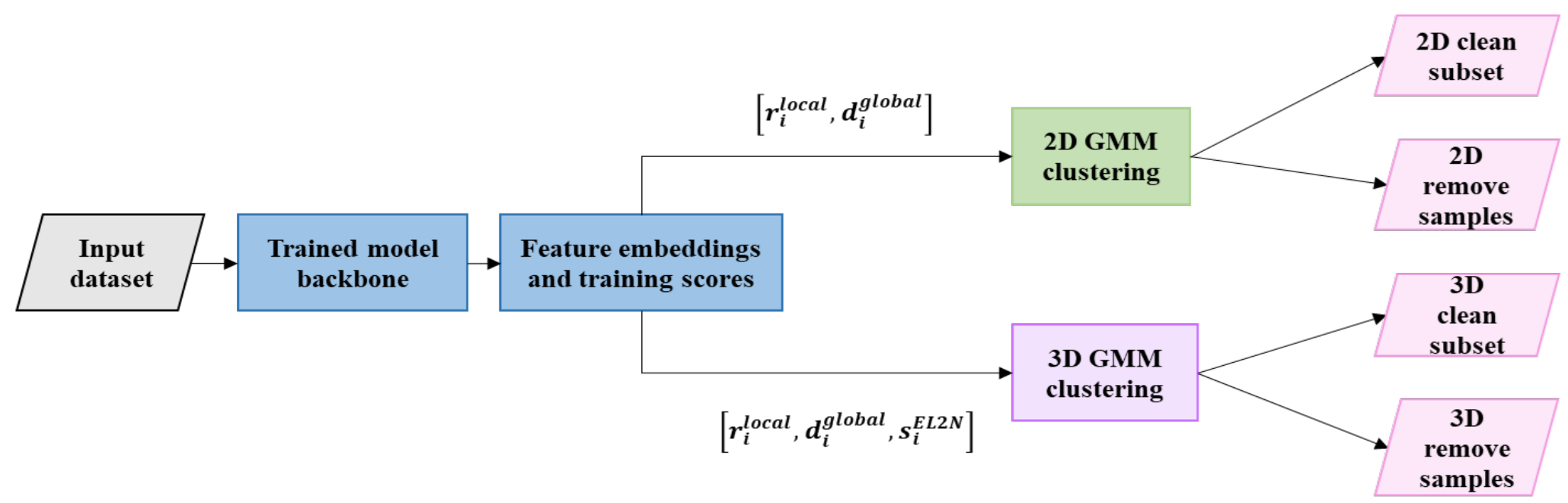

**Figure 1** Overall pipeline of the proposed adaptive data-cleaning framework.

### 3.1 Local Disagreement Ratio

To quantify the local label consistency in the deep feature space, we employ a class-adaptive K-Nearest Neighbors (KNN) strategy and operationalize inconsistency as the local disagreement ratio. In contrast to conventional methods that utilize a fixed neighborhood size $K$, which may fail to capture the local density variations in imbalanced datasets, our approach determines the neighborhood size $K_i$ dynamically according to the population of the observed class $y_i$. Specifically, the neighborhood size is determined by the number of samples in the observed class, allowing the local consistency measurement to adapt to class-size variation while maintaining sublinear growth, we define $K_i$ as:

$$K_i = \sqrt{N_{y_i}} \tag{1}$$

where $N_{y_i}$ denotes the total number of training samples associated with class $y_i$. Based on this adaptive neighborhood, the Local Disagreement Ratio $r_i^{local}$ is computed as the proportion of neighbors that do not share the same observed label:

$$r_i^{local} = 1 - \frac{1}{K_i} \sum_{x_j \in \mathcal{N}_{\mathcal{K}_i}(x_i)} I(y_j = y_i) \tag{2}$$

where $\mathcal{N}_{\mathcal{K}_i}(x_i)$ represents the set of $K_i$ nearest neighbors of $x_i$ determined by cosine similarity, and $I(\cdot)$ is the indicator function. A higher value of $r_i^{local}$ indicates a greater degree of label disagreement between a sample and its local geometric neighborhood, suggesting a higher likelihood of label corruption.

### 3.2 Global Centroid Distance

To capture structural deviations that extend beyond local neighborhoods, we introduce a global centroid-based feature. We first apply $k$-means clustering to the feature embeddings across the entire dataset, with the number of clusters set to the total number of classes to obtain a coarse approximation of the global data structure. This process is strictly unsupervised; each sample $x_i$ is assigned to its nearest centroid $c_{k_i}$ based solely on its position in the feature space, independent of its observed label. To measure the discrepancy between a sample and its assigned global centroid, we compute the cosine distance between the sample and its nearest $k$-means centroid:

$$d_i^{global} = 1 - \frac{x_i^{\top} c_{k_i}}{|x_i||c_{k_i}|} \tag{3}$$

A larger value of $d_i^{global}$ indicates that the sample is farther from its assigned global centroid in the learned feature space. Such structural deviation may be associated with potential label corruption or ambiguous visual content.

### 3.3 Robust EL2N Score

To incorporate the learning dynamics of the model, we utilize the Error L2-Norm (EL2N), which measures the L2 distance between the model's predicted probability vector and the one-hot encoded observed label. Specifically, for a given sample $i$, the raw EL2N score $e_i$ is computed as:

$$e_i = \|P_i - y_i\|_2 \tag{4}$$

where $P_i$ represents the model's predicted probability vector (typically the softmax output), and $y_i$ denotes the one-hot encoded vector of the observed label.

While raw EL2N scores provide a snapshot of learning difficulty, they are sensitive to extreme outliers and varying convergence speeds across classes. We therefore apply a class-conditional robust Z-score transformation. Let $e_{y_i}$ be the set of raw EL2N scores for all samples with observed label $y_i$. For a sample with score $e_i$, the Robust EL2N Score $s_i^{EL2N}$ is defined as:

$$s_i^{EL2N} = 0.6745 \cdot \frac{e_i - \text{median}(e_{y_i})}{\text{MAD}(e_{y_i})} \tag{5}$$

where $MAD(\cdot)$ denotes the Median Absolute Deviation, and 0.6745 is the third quartile of the standard normal distribution [25], which acts as a scaling factor to make MAD a consistent estimator for the standard deviation. MAD is mathematically formulated as the median of the absolute deviations from the dataset's median:

$$\text{MAD}(e_{y_i}) = \text{median}(|e - \text{median}(e_{y_i})|) \tag{6}$$

This transformation effectively standardizes the learning dynamics metric across different classes, yielding a metric that is highly robust against outliers.

### 3.4 Multi-Metric and Adaptive GMM Data Cleaning

The proposed framework adopts a modular strategy that integrates the above indicators into a compact low-dimensional feature space. Two methods variants are considered in this study. The first variant uses only geometric information and constructs a 2D feature space from local disagreement ratio and global centroid distance. The second variant extends this method by incorporating the robust EL2N score, forming a 3D feature space that additionally reflects learning dynamics. These two variants are evaluated as practical alternatives under the same detection and retraining pipeline.

**2D geometric metric:** The framework focuses exclusively on the structural and local geometric consistency of the feature space. Each sample is represented by:

$$z_i^{(2D)} = \left[r_i^{local}, d_i^{global}\right] \quad (7)$$

**3D multi-metric:** The framework extends the geometric metric by incorporating the learning dynamics related EL2N metric to provide a more comprehensive view of the learning dynamics:

$$z_i^{(3D)} = \left[r_i^{local}, d_i^{global}, s_i^{EL2N}\right] \quad (8)$$

Given the different numerical scales of these indicators, we apply standard scaling before fitting a two-component Gaussian Mixture Model (GMM) to the feature space. The GMM performs an unsupervised binary partition to isolate the noise-dominant component $C_{noise}$. The identification of $C_{noise}$ depends on the chosen mode.

For the 2D feature space, since noisy samples typically exhibit higher structural deviation, the component with the higher mean along the Local disagreement dimension is identified as noise:

$$C_{noise}^{2D} = \underset{c \in \{1,2\}}{argmax}\, \mu_c^{local} \quad (9)$$

For the 3D feature space, the component with the higher mean along the Robust EL2N dimension is selected:

$$C_{noise}^{3D} = \underset{c \in \{1,2\}}{argmax}\, \mu_c^{EL2N} \quad (10)$$

Finally, samples assigned to $C_{noise}$ under the GMM posterior partition are removed from the training set. Although this binary decomposition may not fully capture all ambiguous sample types, it provides a simple and adaptive mechanism for isolating dominant label noise without requiring dataset-specific cutoff selection.

# 4 Experimental results

## 4.1 Experimental Setup and Datasets

### 4.1.1 COMPUTING SETUP

All experiments were conducted on a workstation equipped with an Intel Core i7-11700 CPU, 32 GB DDR4 RAM, and an NVIDIA GeForce RTX 3090 GPU. The software environment consisted of Windows 11 24H2, Python 3.10.19, PyTorch 2.7.1+cu118, CUDA 11.8, and cuDNN 9.1.0.

### 4.1.2 Implementation Details

To extract deep feature spaces, we employed ResNet50 and Vision Transformer (ViT) architectures initialized with pre-trained weights [26],[27]. Specifically, the ViT backbone utilized the vit_base_patch16_224 pre-trained configuration. To satisfy the input dimensional requirements of these models and ensure spatial consistency, all original images across the evaluated datasets were uniformly resized to 224×224 pixels prior to training.

#### 4.1.3 Synthetic Label Noise Generation:

To rigorously evaluate the robustness of the proposed framework under varying corruption levels, we synthetically introduced label noise into the datasets at ratios of 5%, 10%, 20%, and 40%. The corruption was implemented following a symmetric noise injection strategy for a given noise ratio, we randomly sampled the corresponding percentage of instances from each individual class and uniformly reassigned their labels to all other available classes. This procedure ensures a controlled and unbiased noise distribution for our empirical validation.

#### 4.1.4 Datasets

**CIFAR-10:**

The CIFAR-10 dataset is a widely adopted benchmark for image classification, comprising 60,000 color images distributed equally across 10 mutually exclusive classes (such as airplanes, automobiles, birds, and cats) [28]. The original spatial resolution of these images is 32×32 pixels. In our experiments, we follow the standard evaluation protocol, utilizing exactly 50,000 images for the training set and reserving the remaining 10,000 images for testing. This diverse, real-world object dataset serves as a standard baseline to evaluate the generalizability of our noise-detection framework across complex feature spaces.

**MNIST:**

The MNIST dataset is a fundamental machine learning benchmark consisting of grayscale images of handwritten digits ranging from 0 to 9 [29]. The original spatial resolution of the images is 28×28 pixels. For our empirical evaluation, we employ the standard dataset split, which provides exactly 60,000 training samples and 10,000 testing samples. Given its relatively simple and highly separable feature space, MNIST serves as an ideal control dataset to validate the framework's baseline efficacy and its adaptive behavior under various noise conditions.

**ImageNet100:**

To further evaluate the scalability and robustness of our framework on more complex, high-resolution data, we utilize the ImageNet-100 dataset [30]. ImageNet-100 is a widely adopted 100-class subset of the large-scale ILSVRC-2012 benchmark, featuring a diverse array of real-world objects with highly complex backgrounds. In our empirical evaluation, we follow the standard subset split, which provides approximately 130,000 training samples and 5,000 testing samples. Compared to other dataset, ImageNet-100 presents a significantly more challenging feature space, serving as a rigorous testbed to validate the framework's adaptability and detection efficacy in large-scale scenarios.

## 4.2 Baselines and Evaluation Metrics

### 4.2.1 Cleanlab (Confident Learning):

To provide a representative comparison with an established probabilistic label-cleaning method, we employ Cleanlab, a widely used open-source implementation of Confident Learning, as the primary comparative baseline. Confident Learning estimates label issues from noisy labels and out-of-sample predicted probabilities, which are typically obtained via cross-validation or other holdout prediction strategies. By operating on model outputs rather than modifying a specific architecture or loss function, Confident Learning decouples the label-cleaning procedure from the final training model; any classifier capable of producing class-probability estimates can be used to generate the required predictions, and the cleaned dataset can subsequently be used to train another model.

In large-scale settings, however, obtaining out-of-sample predictions with high-capacity target architectures such as ResNet50 or ViT can be computationally expensive, especially when k-fold cross-validation is used. Therefore, we implement Cleanlab under a practical proxy-model setting: a lightweight backbone is first used to generate out-of-sample probabilities for label-issue detection, and the resulting cleaned dataset is then used for downstream training with the target architecture. This setting reflects the practical advantage of Confident Learning as a model-agnostic data-cleaning strategy, while also highlighting the computational cost associated with producing reliable out-of-sample predictions.

In contrast, our proposed framework avoids iteration-heavy cross-validation during the noise-detection stage. Instead of relying on a lightweight proxy model, it directly leverages the target backbone used in downstream training. Specifically, our method combines learning dynamics information with geometric consistency in the learned feature space, allowing noise detection to be conditioned of the target model itself. This design enables our framework to exploit the discriminative feature extraction capability of modern deep architectures such as ResNet50 and ViT during the initial data-cleaning phase.

Moreover, rather than using manually specified filtering thresholds or requiring prior knowledge of the noise ratio, our framework employs a distribution-based partitioning strategy over multiple noise-related scores. As a result, the clean/noisy separation can adapt to the score distribution induced by each target backbone and corruption setting. This target-model-driven design provides an efficient and streamlined alternative to proxy-based cleaning, particularly when the objective is to improve the performance of a specific downstream architecture.

#### 4.2.2 Evaluation Metrics

We evaluate the proposed framework from two perspectives. First, noisy-label detection efficacy is quantified using accuracy, precision, and recall, where synthetically corrupted samples are regarded as positive instances and clean samples are regarded as negative instances. The detection results of the proposed 2D geometric metric and 3D multi-metric methods are reported in Tables I–III across CIFAR-10, MNIST, and ImageNet-100, while the Cleanlab baseline is reported in Table IV.

Second, downstream retraining performance is evaluated by removing the samples predicted as noisy, retraining the target backbone on the retained subset, and measuring the resulting test accuracy. In Fig. 6, retraining performance is presented as the accuracy improvement after data cleaning, defined as the percentage-point change in test accuracy relative to training on the original noisy dataset before cleaning.

### 4.3 Detection Efficacy

#### 4.3.1 Proposed Framework Performance

Tables I–III summarize the noisy-label detection performance of the proposed 2D geometric metric and 3D multi-metric across CIFAR-10, MNIST, and ImageNet-100. Noisy samples generated by the synthetic corruption process are treated as positive instances. Overall, the proposed methods maintain consistently high recall across most datasets and corruption levels, indicating that the majority of corrupted labels can be successfully isolated before retraining.

On CIFAR-10, recall is generally above 90% across backbones and noise ratios, while MNIST reaches 99.40%–100% recall in all proposed configurations. ImageNet-100 presents a more challenging feature space: at 5% and 10% noise, precision is relatively low because noisy and hard clean samples overlap, but recall remains high in most settings and reaches roughly 98% at 40% noise. This suggests that the proposed framework prioritizes the removal of potentially unreliable samples, which can be beneficial when the objective is to construct a cleaner subset for downstream retraining.

#### 4.3.2 Comparison with Cleanlab

Table IV reports the detection efficacy of Cleanlab under the proxy-model setting described in the experimental protocol. Cleanlab achieves strong performance on MNIST, where precision ranges from 95.52% to 99.13% and recall ranges from 96.86% to 98.62%.

However, on CIFAR-10 and ImageNet-100, its recall and precision vary more substantially across noise ratios. ImageNet-100 recall remains within 83.43%–86.19%. Compared with this probability-based baseline, the proposed framework directly exploits the target backbone and integrates local, global, and learning dynamics cues, leading to more robust noise isolation in several high-corruption settings.

TABLE I

DETECTION EFFICACY ON CIFAR-10

| Detection Efficacy | | **Resnet50** | | | | **Vit** | | | |
|---|---|---|---|---|---|---|---|---|---|
| Label error(%) | | **5%** | **10%** | **20%** | **40%** | **5%** | **10%** | **20%** | **40%** |
| 2D geometric metric | Accuracy | 94.18% | 95.38% | 95.65% | 96.30% | 90.44% | 92.73% | 91.93% | 89.39% |
| | Precision | 46.10% | 69.40% | 85.33% | 95.82% | 34.02% | 58.37% | 71.74% | 83.85% |
| | Recall | 96.84% | 96.24% | 94.51% | 94.91% | 97.12% | 96.39% | 98.40% | 91.00% |
| 3D multi-metric | Accuracy | 90.53% | 93.44% | 96.12% | 96.32% | 88.00% | 98.74% | 91.12% | 89.66% |
| | Precision | 34.48% | 60.62% | 87.51% | 95.84% | 29.22% | 49.10% | 69.57% | 84.62% |
| | Recall | 99.40% | 98.10% | 94.03% | 94.92% | 98.40% | 89.64% | 98.86% | 90.61% |

TABLE II

DETECTION EFFICACY ON MNIST

| Detection Efficacy | | **Resnet50** | | | | **Vit** | | | |
|---|---|---|---|---|---|---|---|---|---|
| Label error(%) | | **5%** | **10%** | **20%** | **40%** | **5%** | **10%** | **20%** | **40%** |
| 2D geometric metric | Accuracy | 96.55% | 99.83% | 99.77% | 99.69% | 92.70% | 92.42% | 99.49% | 99.28% |
| | Precision | 59.16% | 98.89% | 99.44% | 99.58% | 40.61% | 56.86% | 97.77% | 98.52% |
| | Recall | 99.83% | 99.45% | 99.40% | 99.64% | 99.97% | 99.98% | 99.72% | 99.70% |
| 3D multi-metric | Accuracy | 93.20% | 99.82% | 99.77% | 99.69% | 89.32% | 88.91% | 90.20% | 96.55% |
| | Precision | 42.34% | 98.63% | 99.42% | 99.58% | 31.85% | 47.39% | 67.12% | 92.07% |
| | Recall | 99.97% | 99.57% | 99.42% | 99.65% | 99.97% | 100% | 100% | 99.97% |

TABLE III

DETECTION EFFICACY ON IMAGENET100

| Detection Efficacy | | **Resnet50** | | | | **Vit** | | | |
|---|---|---|---|---|---|---|---|---|---|
| Label error(%) | | **5%** | **10%** | **20%** | **40%** | **5%** | **10%** | **20%** | **40%** |
| 2D geometric metric | Accuracy | 80.47% | 81.67% | 88.35% | 97.50% | 77.60% | 77.79% | 90.01% | 96.03% |
| | Precision | 20.31% | 35.29% | 63.25% | 95.75% | 18.06% | 31.01% | 67.14% | 92.64% |
| | Recall | 99.37% | 99.87% | 99.71% | 98.11% | 98.40% | 99.66% | 68.03% | 97.86% |
| 3D multi-metric | Accuracy | 80.21% | 83.81% | 97.98% | 97.69% | 75.72% | 79.02% | 92.71% | 96.18% |
| | Precision | 20.12% | 38.17% | 91.94% | 95.70% | 17.02% | 32.25% | 74.15% | 92.62% |
| | Recall | 99.62% | 99.83% | 98.55% | 98.65% | 99.46% | 99.75% | 97.56% | 98.29% |

TABLE IV

DETECTION EFFICACY OF THE CLEANLAB BASELINE

| Label error(%) | | **5%** | **10%** | **20%** | **40%** |
|---|---|---|---|---|---|
| **CIFAR-10** | Accuracy | 95.51% | 93.76% | 89.55% | 82.34% |
| | Precision | 52.87% | 62.50% | 67.26% | 72.38% |
| | Recall | 93.72% | 93.96% | 93.07% | 90.31% |
| **MNIST** | Accuracy | 99.62% | 99.57% | 99.35% | 99.10% |
| | Precision | 95.52% | 97.66% | 98.57% | 99.13% |
| | Recall | 96.86% | 98.07% | 98.18% | 98.62% |
| **ImageNet100** | Accuracy | 86.92% | 85.60% | 84.45% | 81.32% |
| | Precision | 25.41% | 39.70% | 57.41% | 72.87% |
| | Recall | 83.43% | 84.82% | 86.19% | 84.90% |

### 4.4 Qualitative Analysis of GMM Partitioning

To visually demonstrate the effectiveness of the proposed feature representations, the visualizations in Fig. 2 to Fig. 5 are generated based on the MNIST dataset using the ResNet50 backbone under 10% symmetric label noise. In each figure, the ground-truth clean/noisy labels are used only for visualization and evaluation, whereas the GMM partition is obtained without accessing the true corruption labels.

#### 4.4.1 2D Geometric Space

Fig. 2 shows the distribution of samples in the 2D geometric metric space formed by the local disagreement ratio and global centroid distance. As shown in Fig. 2(a), the synthetically corrupted samples tend to appear in regions with higher local disagreement and larger centroid distance, while most clean samples are concentrated in the low-disagreement region. Fig. 2(b) shows that the two-component GMM captures a similar high-risk region and assigns it to the noisy component. This result suggests that combining local and global geometric cues provides a distinguishable structure for adaptive noisy-label detection.

Fig. 3 further presents the t-SNE visualization of the same 2D metric. The ground-truth distribution in Fig. 3(a) and the GMM-predicted partition in Fig. 3(b) show similar clean/noisy separation patterns, indicating that the engineered 2D metric preserves a meaningful structure after nonlinear projection.

### 4.4.2 3D Multi-metric Space

Fig. 4 visualizes the 3D multi-metric space, which combines local disagreement ratio, global centroid distance, and robust EL2N score. Compared with the 2D metric, the additional robust EL2N dimension introduces learning-dynamics information into the feature representation. In Fig. 4(a), noisy samples form a distinguishable region in the fused multi-metric space. Fig. 4(b) shows that the GMM-predicted noisy component is largely aligned with this high-risk region. Fig. 5 shows the t-SNE projection of the 3D multi-metric representation. The comparison between Fig. 5(a) and Fig. 5(b) further demonstrates that the GMM-predicted noisy component follows a distribution pattern similar to the ground-truth corrupted samples, supporting the effectiveness of incorporating robust EL2N with geometric metrics.

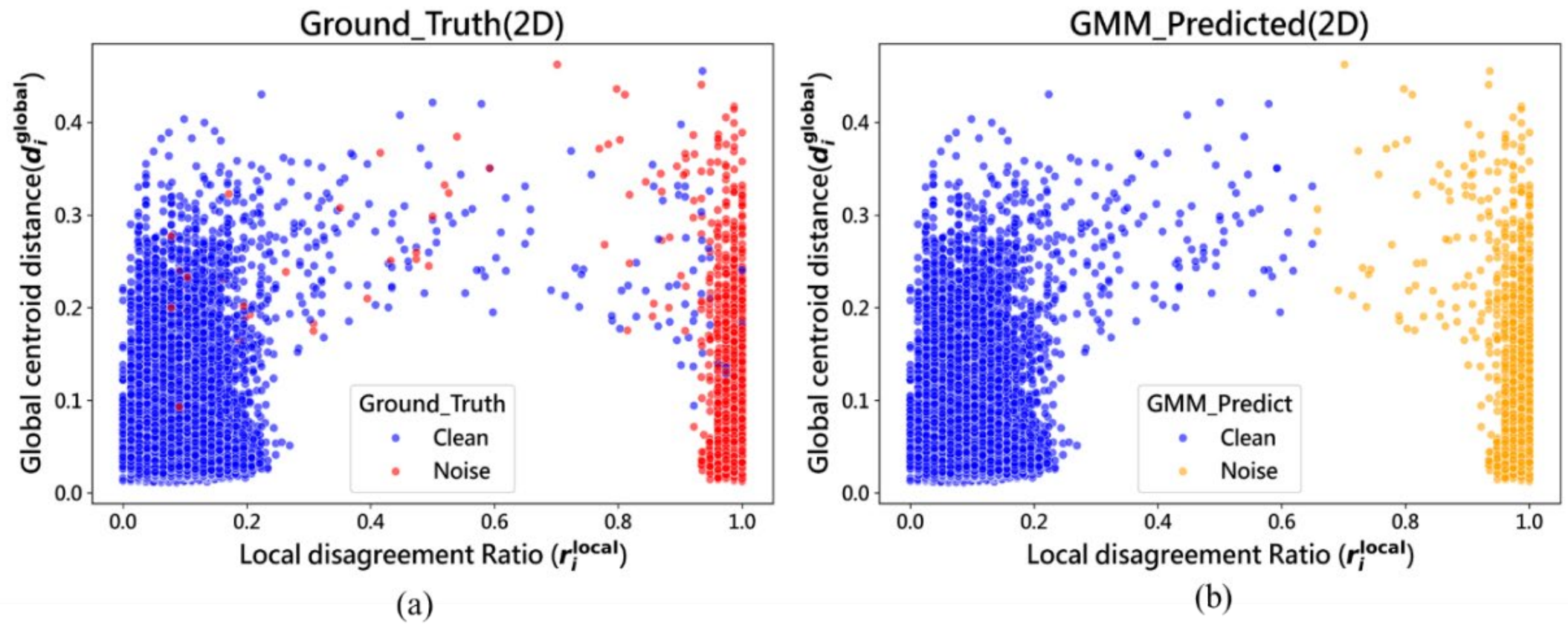

**Figure 2** 2D GMM partitioning in the geometric metric space. (a) Ground-truth clean/noisy labels. (b) GMM-predicted clean/noisy partition.

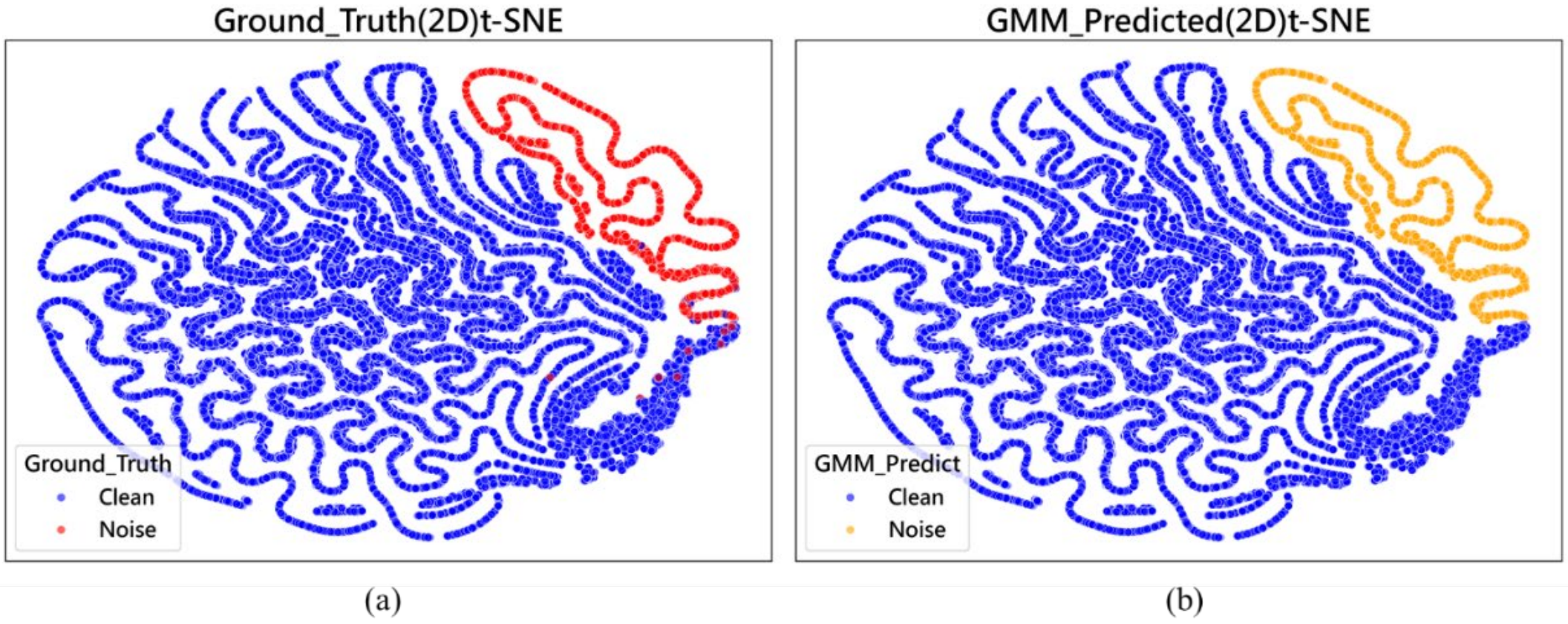

**Figure 3** t-SNE visualization of the 2D geometric metric. (a) Ground-truth clean/noisy labels. (b) GMM-predicted clean/noisy partition. structure.

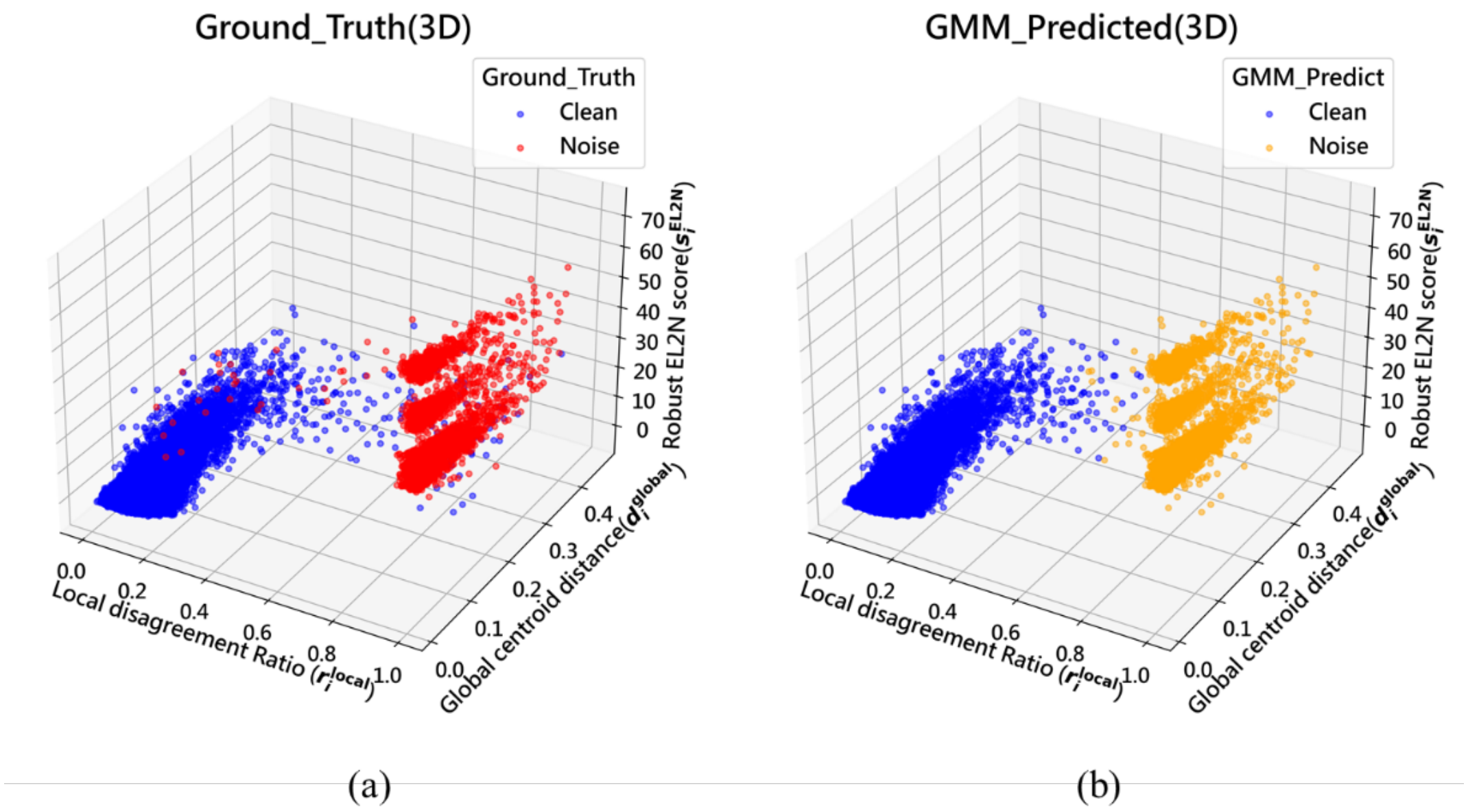

**Figure 4** 3D GMM partitioning in the multi-metric space. (a) Ground-truth clean/noisy labels. (b) GMM-predicted clean/noisy partition.

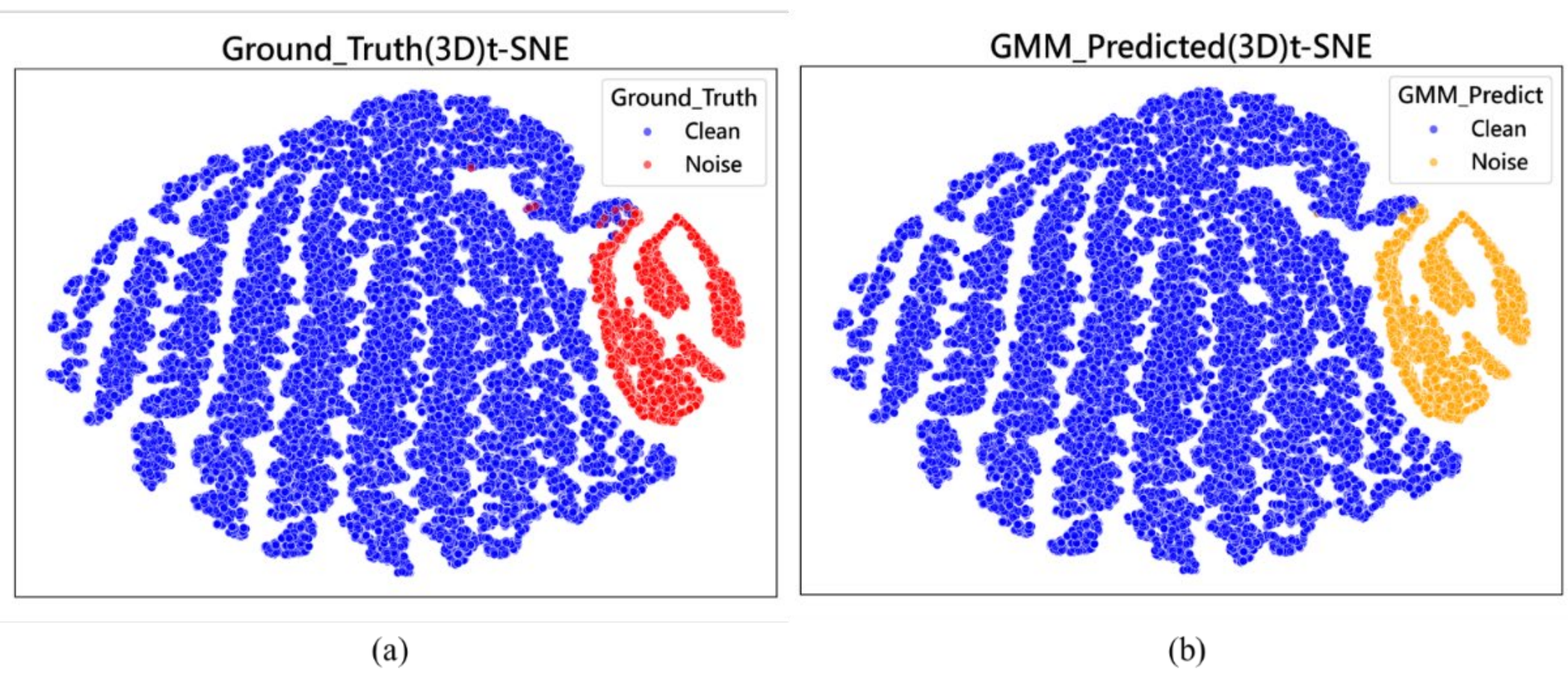

**Figure 5** t-SNE visualization of the 3D multi-metric representation. (a) Ground-truth clean/noisy labels. (b) GMM-predicted clean/noisy partition.

## 4.5 Retraining Performance

Fig. 6 compares the downstream retraining performance after applying different data-cleaning strategies, with the results for CIFAR-10, MNIST, and ImageNet-100 shown in Fig. 6(a), Fig. 6(b), and Fig. 6(c), respectively. The y-axis represents the percentage-point change in test accuracy relative to training on the original noisy dataset before cleaning. Overall, the proposed 2D geometric metric and 3D multi-metric methods improve downstream accuracy in many evaluated settings, especially under moderate and severe label-noise conditions. This result indicates that removing the samples identified as noise-dominant by the adaptive GMM partition can effectively reduce the negative influence of corrupted labels during subsequent model training.

On CIFAR-10, the proposed methods generally provide positive accuracy gains when the noise ratio increases, particularly under the ViT backbone at 20% and 40% label noise. This suggests that, when the feature becomes sufficiently discriminative, the proposed local-global geometric cues and the additional EL2N-based learning dynamics cue can identify noisy or highly unreliable samples that would otherwise degrade generalization. In contrast, Cleanlab exhibits less stable retraining behavior in several CIFAR-10 settings, especially under the ViT backbone, where the proxy-model-based cleaning result does not always translate into improved downstream performance. This comparison suggests that directly conditioning the cleaning process on the target backbone can be advantageous when the objective is to improve a specific downstream architecture.

On MNIST, all methods produce relatively small accuracy changes. This is mainly because MNIST is a simpler and highly separable dataset, and the baseline accuracy before cleaning is already close to saturation. Consequently, even when noisy labels are successfully detected, the available margin for further improvement is limited. In this setting, Cleanlab remains competitive because its probability-based label-issue estimation performs well on highly separable data, whereas the proposed methods achieve comparable retraining behavior with only minor fluctuations.

The advantage of the proposed framework becomes more evident on ImageNet-100, which presents a more complex feature space and stronger inter-class ambiguity. Under severe corruption, especially at 40% label noise, the proposed 3D multi-metric method with ViT achieves the largest improvement among the compared methods in Fig. 6. Although Cleanlab also improves retraining performance under high-noise ImageNet-100 settings, its gain is smaller than that of the proposed 3D multi-metric method. This result indicates that integrating local disagreement, global centroid distance, and robust EL2N score provides a more effective refinement signal for complex datasets, where probability estimates from a lightweight proxy model may not fully capture the noise structure relevant to the target backbone.

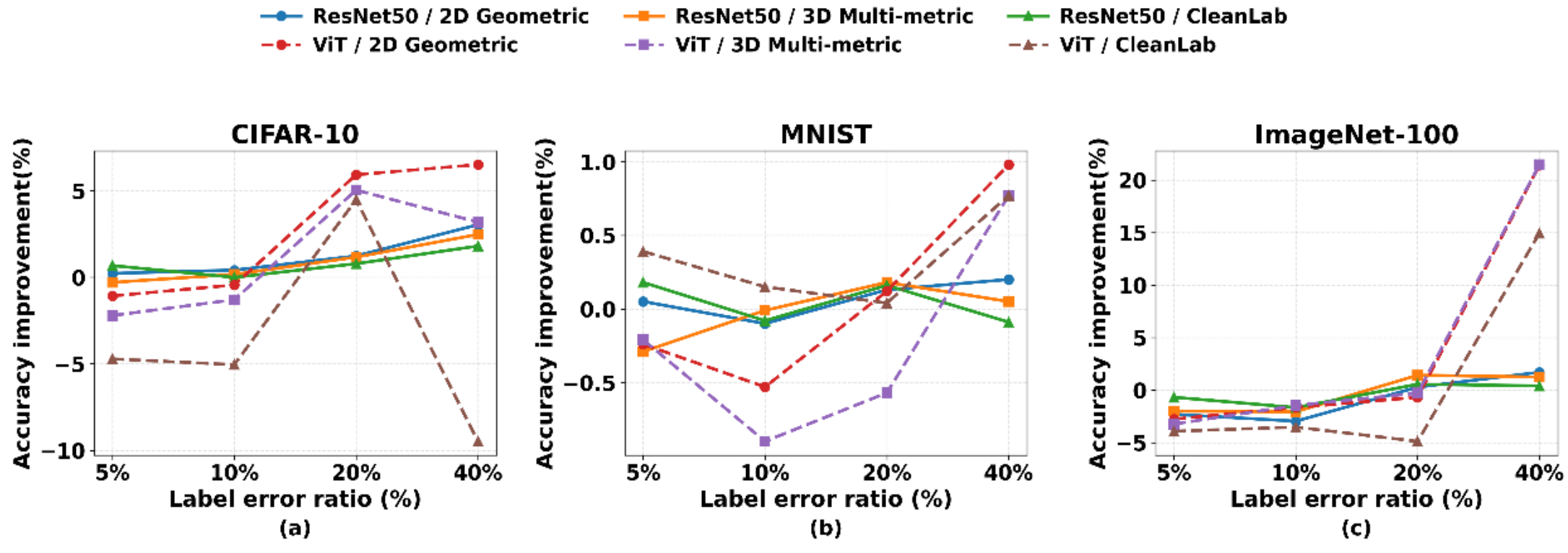

**Figure 6** Accuracy improvement after data cleaning. (a) CIFAR-10. (b) MNIST. (c) ImageNet-100.

Negative improvements are observed in several low-noise or highly overlapping settings. These cases indicate that aggressive filtering may remove informative hard samples together with mislabeled samples, or reduce the representativeness of the retained subset. Therefore, the retraining results should be interpreted together with the detection results: the proposed framework prioritizes high-recall noise isolation and tends to construct a cleaner but sometimes more conservative training subset. Overall, Fig. 6 demonstrates that the proposed adaptive data-cleaning strategy is particularly beneficial when label corruption is severe and when the dataset contains complex visual structures.

# 5 Discussion and Conclusion

## 5.1 Low-Noise Regime: Separability Rather Than Noise Ratio Alone

In low-noise settings, the proposed framework does not necessarily exhibit weaker noisy-label detection performance. Instead, its behavior is strongly influenced by the separability of the learned feature and the interaction between the dataset and backbone architecture. When clean and corrupted samples are not clearly separated in the feature space, the GMM partition may isolate not only mislabeled instances but also structurally ambiguous yet correctly labeled hard samples. This phenomenon can lead to relatively conservative precision in strict noisy-label detection, while still improving retraining performance by removing confusing boundary cases and producing a cleaner, more stable training subset.

Importantly, this conservative precision is accompanied by high recall rates across all evaluated noise ratios. As demonstrated in our experiments on each dataset, the recall performance of both 2D and 3D metric methods consistently remains outstanding, frequently exceeding 95% and reaching near 100% in many cases. This high recall ensures that the vast majority of genuine noisy labels are successfully identified and intercepted. Consequently, while the framework may aggressively prune some structurally ambiguous clean samples, it effectively minimizes the risk of missing hidden label corruption, thereby reducing the likelihood that noisy labels remain in the retained subset that drives the subsequent improvement in retraining performance. Therefore, the proposed framework should be interpreted as an adaptive data-refinement strategy rather than a strict label-error detector that perfectly separates clean and noisy instances.

However, the experimental results also show that low-noise conditions can yield excellent detection performance when the feature space is sufficiently discriminative.

For example, on MNIST with ResNet50, both the 2D geometric metric and 3D multi-metric achieve near-perfect detection even under 10% label noise. This indicates that low noise ratio itself is not the direct cause of weaker detection. Rather, detection difficulty depends on whether noisy samples form a distinguishable distribution relative to clean samples.

Therefore, the role of the proposed framework in low-noise regimes should be interpreted as adaptive data refinement: in challenging feature spaces, it may filter both noisy and ambiguous hard samples, whereas in highly separable feature spaces, it can accurately isolate label errors with minimal confusion.

### 5.2 High-Noise Regime: Stable Noise Isolation Through Multi-Metric

Under high-noise conditions, the proposed framework generally demonstrates strong robustness because corrupted samples become more statistically visible in the feature space. As the noise ratio increases, mislabeled instances are more likely to form a sufficiently large and coherent noise-dominant component, allowing the Gaussian Mixture Model to estimate the corresponding distribution more stably. This explains the strong detection results observed in several high-noise experiments, and suggests that multi-metric can provide useful evidence for noisy-label detection when confidence-based estimates alone may become less reliable.

This robustness is further supported by the integration of complementary metrics. Unlike methods that rely primarily on predicted probabilities, the proposed framework combines local neighborhood disagreement, global centroid distance, and, in the 3D variant, robust EL2N-based learning dynamics. As a result, even when model confidence becomes unreliable under severe label corruption, the geometric structure of the feature space can still provide useful evidence for noisy-label detection. The 2D and 3D variants also exhibit complementary behavior: the 2D metric may provide greater stability when learning dynamics are noisy or unstable, whereas the 3D metric can offer finer discrimination when the EL2N metric is reliable.

### 5.3 Overall Implications

Overall, the experimental findings suggest that the effectiveness of noisy-label detection is not determined by the label-noise ratio alone. Instead, it depends on the joint effect of noise ratio, dataset complexity, backbone quality, and the separability of clean and noisy samples in the feature space. The proposed multi-dimensional GMM framework is therefore best characterized as an adaptive, threshold-free data refinement mechanism. In highly separable settings, such as MNIST with ResNet50, it can accurately detect noisy labels even at low corruption levels. In more complex or overlapping feature spaces, especially under low-noise conditions, it may also remove ambiguous hard samples that are not necessarily mislabeled but can still hinder generalization.

A defining strength of this adaptive refinement mechanism is its capacity to maintain consistently high recall across varying corruption levels. Empirical results indicate that across different datasets and noise ratios (e.g.,5%, 10%, 20%, and 40%), the proposed framework achieves remarkable recall rates, effectively preventing noisy labels from slipping through the detection process. By heavily penalizing potential noise and prioritizing high recall, the framework establishes a cleaner training subset. This extensive noise isolation fundamentally underpins the robust performance recovery observed in our retraining experiments, proving that removing ambiguous instances alongside genuine noise is a favorable trade-off for overall model robustness. Under severe noise, the multi-metric provides robust anchors for isolating dominant noisy components and mitigating performance degradation.

Furthermore, unlike traditional probabilistic methods that rely on computationally expensive k-fold cross-validation to obtain out-of-sample predictions, the proposed framework directly leverages the target backbone without requiring iteration-heavy cross-validation. This architectural advantage significantly reduces the overall computational overhead and training costs, making it highly efficient for large-scale applications. These results highlight the practical value of the proposed framework: it does not rely on a fixed assumption that low-noise or high-noise regimes are inherently easier or harder. Instead, it adapts to the empirical structure of the learned, enabling flexible noisy-label detection and data cleaning across different datasets, architectures, and corruption levels.